\documentclass[
]{ceurart}

\usepackage{listings}
\begin{document}

\copyrightyear{2022}
\copyrightclause{Copyright for this paper by its authors.
  Use permitted under Creative Commons License Attribution 4.0
  International (CC BY 4.0).}

\conference{FDIA 2022: Symposium on Future Directions in Information Access (FDIA), July 07-20, 2022, Lisbon, Portugal.}

\title{Building a Relation Extraction Baseline for Gene-Disease Associations: A Reproducibility Study}

\author[1]{Laura Menotti}[%
email=laura.menotti@studenti.unipd.it,
]
\address[1]{Department of Information Engineering, University of Padua, Padua, Italy}

\begin{abstract}
  Reproducibility is an important task in scientific research. It is crucial for researchers to compare newly developed systems with the state-of-the-art to assess whether they made a breakthrough. However previous works may not be immediately reproducible, for example due to the lack of source code. In this work we reproduce DEXTER, a system to automatically extract Gene-Disease Associations (GDAs) from biomedical abstracts.\cite{dexter} The goal is to provide a benchmark for future works regarding Relation Extraction (RE), enabling researchers to test and compare their results.
\end{abstract}


\maketitle

\section{Introduction}
Biomedical literature is a rich source of information on Gene-Disease Associations (GDAs) that could help physicians in assessing clinical decisions and improve patient care. GDAs are publicly available in databases containing relationships between gene/miRNA expression and related diseases such as specific types of cancer. Most of these resources, such as DisGeNET\cite{disgenet1, disgenet2}, miR2Disease\cite{mir2disease} and BioXpress\cite{bioxpress1, bioxpress2}, include also manually curated data from publications. Human annotations are expensive and cannot scale to the huge amount of data available in scientific literature (e.g., biomedical abstracts). Therefore, developing automated tools to identify GDAs is getting traction in the community. Such systems employ Relation Extraction (RE) techniques to extract information on gene/microRNA expression in diseases from text. Once an automated text-mining tool has been developed, it can be tested on human annotated data or it can be compared to state-of-the-art systems. 

In this context, the state of the art is DEXTER, a rule-based system that extracts gene/microRNA expressions in diseases from biomedical abstracts\cite{dexter}. Unfortunately, DEXTER's source code is not publicly available, thus it cannot be easily used as a baseline for relation extraction experiments. In this work we reproduce DEXTER to provide a benchmark for RE, thus enabling researchers to test and compare their results to a state-of-the-art baseline.

The rest of the paper is structured as follows: Section 2 introduces DEXTER and its main components, Section 3 describes our implementation of DEXTER by highlighting the main differences and commonalities with the original system, Section 4 presents the obtained results and Section 5 draws some conclusions. 

The implemented version of DEXTER is available in the following git repository: \url{https://github.com/mntlra/DEXTER}.

\section{Original System}
DEXTER is a system that extract Gene-Disease Associations (GDAs) starting from biomedical abstracts\cite{dexter}. DEXTER has been published in \textit{Database: The Journal of Biological Databases and Curation} and has attracted 11 citations so far. To the best of our knowledge it is the most recent system for RE on Gene-Cancer Associations. One of the main objectives of the original work is to create a tool to automatically extend expression databases like BioXpress \cite{bioxpress1, bioxpress2}, that contains gene/miRNA expression associated with cancer. The authors focus on a specific set of sentences that report differential expression of genes with or without an explicit comparison between tissues or disease states. Such choice can be motivated by the additional constraints needed to extend BioXpress, i.e. the detection of differential expression of genes compared to healthy tissues. This kind of sentences usually fit into two broad categories called TypeA and TypeB sentences.  
In TypeA sentences a gene's expression is typically contrasted between two different samples or conditions. Such sentences are typically comparative as shown in Example 1, where \emph{Nucleolin expression} is analyzed in non-small cell lung carcinoma (NSCLC) tissues compared with normal lung tissues.
\begin{quote}
    Example 1: Nucleolin expression was higher in NSCLC tissues than adjacent normal lung tissues. \textit{(TypeA sentence)}
\end{quote}
On the other hand, TypeB sentences still provide the expression level of a gene in a particular sample, however there is no explicit comparison. We can see an example of this type of sentences in Example 2, where \emph{miR-630} is detected in NSCLC tissuses without any comparison with other samples. 
\begin{quote}
    Example 2: Our results showed that miR-630 was significantly down-regulated in NSCLC tissues and cell lines. \textit{(TypeB sentence)}
\end{quote}
Among biomedical literature regarding gene expression in diseases one can also find a third category called TypeC sentences. In such sentences gene's expression level and related diseases are still reported, however there is no explicit association between the gene and the disease. 
\begin{quote}
    Example 3: Over-expression of C1GALT1 enhanced breast cancer cell growth, migration and invasion \textit{in vitro} as well as tumor growth \textit{in vivo}. \textit{(TypeC sentence)}
\end{quote}
In Example 3, we see that \emph{when} C1GALT1 is over-expressed in breast cancer, cell-growth is enhanced. However, there is no explicit association between the gene and breast cancer, meaning there is no information about whether C1GALT1 is typically over-expressed in breast cancer or not. DEXTER does not extract information from such sentences.

\begin{figure}
  \centering
  \includegraphics[width=\linewidth]{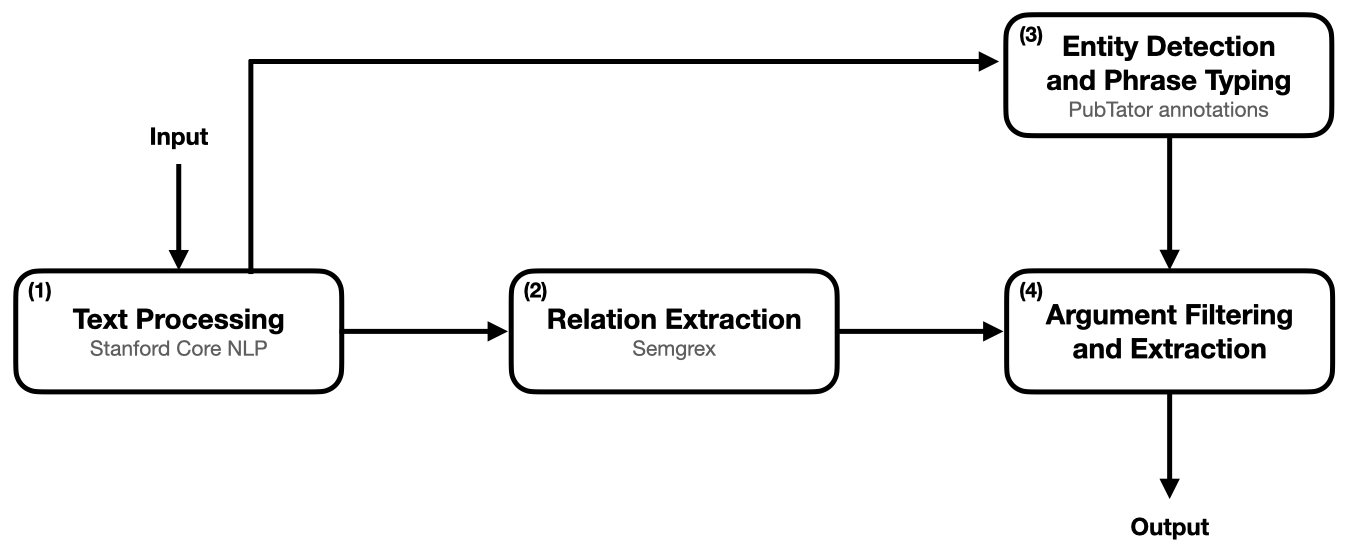}
  \caption{DEXTER modules defined in the original paper. Arrows indicates the direction of the computation, providing an overview of the input of each block.}
  \label{fig:pipeline}
\end{figure}

DEXTER is an expert system, whose rules are based on the syntactic structure of TypeA and TypeB sentences and it extracts information on gene or miRNA expressed in an associated disease. The system is developed mainly in Python and Java and it is based on several modules as shown in Figure \ref{fig:pipeline}, each dealing with a different part of the computation. 

In the text processing block, a given medical abstract is tokenized and split into sentences. Such operation is performed by Stanford CoreNLP toolkit, which is a pipeline that takes raw text as input and returns a set of Natural Language Processing (NLP) annotations as output.\cite{manning-etal-2014-stanford} CoreNLP can be useful in a number of NLP tasks including Part-Of-Speech tagging, Named Entity Recognition and Dependency Parsing. The RE module extracts information from the sentences by using patterns based on lemmas, part-of-speech tags and dependency labels. Therefore using this library is essential for achieving DEXTER's objectives. 
After tokenization and sentence splitting, sentences that do not contain predefined trigger words are discarded and not processed any further. Such terms are available in one of the two supplementary files provided as appendixes of the original paper and they are used to detect sentences that might contain expressions and comparative relations. Each sentence is parsed using the Charniak-Johnson parser \cite{charniak-2000-maximum, charniak-johnson-2005-coarse} with David McClosky's adaptation to the biomedical domain \cite{domain-adaptation} because they provide constituency parse trees that are then converted into Standard Dependency Graphs (SDGs)\cite{de-marneffe-etal-2014-universal}. The resulting SDG is further processed collapsing and propagating dependencies to properly treat sentences involving conjunctions.     

In the RE module, patterns are written in Semgrex\footnote{https://nlp.stanford.edu/nlp/javadoc/javanlp/edu/stanford/nlp/semgraph/semgrex/SemgrexPattern.html}, which is also part of the Stanford NLP toolkit and it allows the matching of nodes and edges in a dependency graph. In particular, nodes in a SDG are the tokens corresponding to sentence words and patterns are written to check a set of attributes for each token. Attributes comprise lemmas, part-of-speech tags and dependency labels between the considered node and already-matched nodes. Authors rely on a previous work where they identify comparative structure in biomedical text\cite{gupta-etal-2017-identifying} to write comparison patterns used to extract the components. All comparison patterns are available in a supplementary file of the original paper. In typeA sentences, the RE module extracts the aspect being expressed (Compared Aspect), the level of expression of such aspect (Scale Indicator) and the entities being compared in the sentence (Compared Entities), that can be two samples of tissue or two disease states. In TypeB sentences the components extracted are basically the same, the only difference is that there is only one entity where the gene/miRNA is expressed. A different terminology is used in the original paper to identify TypeB components such as Expressed Aspect, Expressed Location and Level Indicator. To better understand how this module works we provide some examples both for TypeA and TypeB sentences in Example 4 and the corresponding components extracted by the RE module are shown in Table \ref{tab:components}. For the sake of simplicity, we used the same terminology both for TypeA and TypeB components.
\begin{quote}
    Example 4:
    \begin{enumerate}
        \item[\#1] The expression of \emph{Sam68} was significantly \emph{elevated} in \emph{NSCLC tissues} as compared with \emph{adjacent non-cancerous tissues}.[PMID:24522888]
        \item[\#2] \emph{Lynx1} levels are \emph{decreased} in \emph{lung cancers} compared to \emph{adjacent normal lung}.[PMID:26025503]
        \item[\#3] Expression of \emph{EphA2} is \emph{increased} in \emph{NSCLC metastases}.[PMID:20360610] 
        \item[\#4] We demonstrated that \emph{miR-195} expression was \emph{lower} in \emph{tumor tissues} and was associated with poor survival outcome.[PMID:25840419]
    \end{enumerate}
\end{quote}
In Example 4, the first two sentences are of TypeA since there is an explicit comparison between two entities. In particular, in the first sentence \textit{'Sam68'} expression is compared in \textit{'non-small-cell lung carcinoma (NSCLC)'} and \textit{'non-cancerous tissues'}. Consider sentence \#2, the RE module extracted \textit{'Linx1'} as the gene expressed (i.e., the Compared Aspect), \textit{'decreased'} as the level of expression (i.e., the Scale Indicator) and the entities being compared (i.e., Compared Entity 1 and 2) are \textit{'lung cancers'} and \textit{'adjacent normal lung'} as shown in Table \ref{tab:components}.
On the other hand, sentences \#3 and \#4 are of TypeB since there is only one entity where the gene/miRNA is expressed. In fact, consider sentence \#3, increased expression of \textit{'EphA2'} is detected in \textit{'NSCLC metastasis'} but there is no other entity compared to it. In this case we can see from Table \ref{tab:components} that the RE module extracted \emph{'EphA2'} as the Compared Aspect, \emph{'increased'} as the Scale Indicator and there is only one Compared Entity identified in \emph{'NSCLC metastases'}.
\begin{table}
  \caption{Components extracted from TypeA and TypeB sentences from Example 4. The Compared Aspect is the gene/miRNA being expressed, the Scale Indicator is the level of expression of such aspect and the Compared Entities are the entities being compared in the sentence. Note that in TypeB sentences there is only one entity since the structure of the phrase contain only one tissue or disease state where the gene/miRNA has been detected.}
  \label{tab:components}
  \tabcolsep=0.11cm
  \begin{tabular}{cccccc}
    \toprule
    & Sentence Type & Scale Indicator & Compared Aspect & Compared Entity 1 & Compared Entity 2\\
    \midrule
    \#1 & TypeA & Elevated & Sam68 & NSCLC tissues & non-cancerous tissues\\
    \#2 & TypeA & Decreased & Lynx1 & lung cancers & normal lung\\
    \#3 & TypeB & Increased & EphA2 & NSCLC metastases & \\
    \#4 & TypeB & Lower & miR-195 & tumor tissues & \\
    \bottomrule
  \end{tabular}
\end{table}

The entity detection and phrase typing module takes sentences parsed by the text processing module and determine if there are terms referring to entities of type gene/miRNA, expression or disease/disease-sample. PubTator\cite{pubtator} is an automatic system providing annotations of biomedical concepts such as genes and diseases whose pre-computed annotations are employed to detect genes and diseases mentions in abstracts. On the other hand, microRNA mentions are detected using regular expressions based on the naming convention described in miRBase\cite{mirbase}. Finally, to determine whether a phrase is of type 'Expression' its terms are checked against a list of expression triggers such as 'expression', 'level, 'over-expression', etc. The same strategy has been adopted to detect disease-sample phrase type. In this case triggers can be 'tissues', 'cells', 'patients', 'samples', etc. A full list of such triggers is provided in the supplementary files of the original paper. 

The argument filtering and extraction module consists of two main steps: verify if the tokens matched by the RE module are of the right type, i.e. contains the required mentions, and extract the relevant information such as the gene/miRNA expressed and the associated disease. 
To perform both tasks this module takes the results of the RE module and the mentions identified in the previous block as input. For TypeA sentences, the compared aspect must be of type expression, while the two compared entities need to be of type disease/disease-sample. Similarly, for typeB sentences the expressed aspect and the expression level must be respectively of type expression and disease/disease-sample. If the checks succeed, relevant information are extracted from the results computed by the RE module. This includes the expressed gene, its level and the associated disease. The gene/miRNA expressed is extracted from the compared aspect/expressed aspect argument using the mentions extracted by the entity detection and phrase typing module. If the system detects a gene expression it also returns the corresponding NCBI Gene ID downloaded from PubTator together with the gene mentions. Expression level information is captured by the RE module in the scale indicator/level indicator argument, therefore the level is normalized to high or low by matching the argument against a list of triggers available in the supplementary files. Some examples of triggers for 'high' expression are 'over-expressed', 'increased', etc, while terms like 'low', 'decreased', 'down-regulated' refers to 'low' expression level. Extracting the associated disease requires some extra care. Disease mentions downloaded from PubTator are searched in the compared entities/expression location components, however in some cases the entities contain only generic diseases such as 'tumor', 'cancer' or 'disease'. In these cases, the associated disease is extracted from the abstract by looking for disease mentions in the title or in the first sentence. Alternatively the associated disease can be inferred from sentences in the 'methods' part of the abstract or from sentences describing the experimental set-up. Such sentences are detected by checking if they contain certain 'investigation triggers' such as 'investigated', 'examined', 'analyzed', etc or 'analyzed triggers' such as 'tested', 'collected', 'explored', etc. PubTator returns disease mentions normalized to MEDIC IDs, however such IDs are mapped to DOIDs to allow an easy integration in BioXpress.  

\section{Implemented Version}
\begin{figure}
  \centering
  \includegraphics[width=\linewidth]{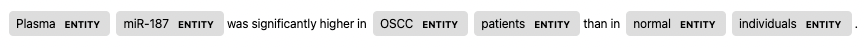}
  \caption{Standard SpaCy pipeline retrieved 6 entities for sentence in Example 5. Entities detected by the standard pipeline are highlighted in grey. The visualization is done using the DisplaCy package.}
  \label{fig:before-exp}
\end{figure}

\begin{figure}
  \centering
  \includegraphics[width=\linewidth]{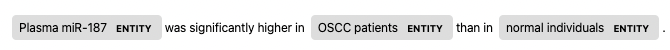}
  \caption{Our custom entity expansion component detected 3 entities for sentence in Example 5, which corresponds to the Compared Aspect and the two Compared Entities. Entities detected by the standard pipeline are highlighted in grey. We can notice how the custom component collapsed the 6 entities detected by the standard pipeline into 3 entities. The visualization is done using the DisplaCy package.}
  \label{fig:after-exp}
\end{figure}

The original work does not share the source code, therefore for the reproducibility we used the description of the paper and the supplementary files containing comparison patterns and trigger lists. As mentioned before, DEXTER is developed using both Python and Java therefore each block run separately from the others. Our goal is to implement an end-to-end system publicly available and easy to reproduce without knowing exactly what each block does. For this reason, we decided to implement DEXTER as an end-to-end application that takes as input biomedical abstracts and returns relevant information on the expressed gene/microRNA and the associated disease. We preserved the overall block architecture and we made some changes in each block to enable a seamless integration of the different modules. The system is entirely developed in Python therefore it was no longer possible to use Stanford CoreNLP toolkit since it is written in Java. CoreNLP developers published Stanza\cite{stanza}, that is a Python NLP package that includes an interface to the CoreNLP Java package, however we decided not to use it since at the time of writing it does not offer support for dependency parsing. Instead we used the SpaCy library\cite{spacy}, which provides annotated text using some trained pipelines available on their website.\footnote{https://spacy.io/usage/models} We used the \texttt{en\textunderscore sci\textunderscore sm} trained pipeline provided by ScispaCy \footnote{https://allenai.github.io/scispacy/}, that is a Python package containing spaCy models for processing biomedical data. In addition, we added a custom component to the SpaCy pipeline to expand entity mentions using hand-craft rules which will be useful when extracting information from the matched components. An example of how entity expansion works is shown in Figures \ref{fig:before-exp} and \ref{fig:after-exp} for the sentence in Example 5. 
\begin{quote}
    Example 5:
    Plasma miR-187 was significantly higher in OSCC patients than in normal individuals.
\end{quote}

In particular, we can notice from Figure \ref{fig:before-exp} and \ref{fig:after-exp} how \emph{OSCC patients} were considered as two separate entities but after the entity expansion they form a single entity. 


Once the system has parsed each sentence we check the presence of miRNA mentions. In that case we tokenize again the sentence to make sure each miRNA mention is considered as a single token. Such operation ensures dependencies are computed correctly and it will be useful in the argument filtering and extraction module.

The RE module is very similar to the one described in the paper\cite{dexter} since the DependencyMatcher library from SpaCy allows to match patterns within the dependency tree of a sentence using Semgrex operators. As a consequence we adapted the comparison patterns available in the supplementary files to match the patterns suitable for the DependencyMatcher library and we added some rules to return matches for more sentences. In fact, we noticed that SpaCy returns different dependency trees than the Stanford CoreNLP library. For this reason we adapted the rules to better handle these cases and to prevent any foreseeable loss of information.

The entity detection and phrase typing module and the argument filtering and extraction module are not significantly different from the original version of DEXTER. Since we downloaded the PubTator annotations in batches we store all the annotations together in such a way that if PubTator failed to detect some mentions from a specific PubMed ID (PMID) we can still retrieve such information if the same gene or disease appears in other abstracts. Furthermore, we added some term triggers for the expression level normalization to cover all possible levels that appear in the sentences. 

\section{Evaluation}
Authors tested DEXTER on data relevant to BioXpress and on a large-scale dataset. To extend the literature-based portion of the BioXpress database they considered three use-cases. Firstly, they evaluated DEXTER on a set of abstracts related to lung cancer. Secondly, they focused on a set of abstracts related to a set of genes, called glycosyltransferases (GTs), which are a set of enzymes. Lastly, authors run the system on a set of abstracts related to the role of microRNA in cancer. The authors tested the system against human annotated data and published the results computed by DEXTER on the website of BioXpress.\footnote{https://hive.biochemistry.gwu.edu/bioxpress}

To evaluate the effectiveness of the reproduced version of DEXTER we run our implementation of the system on the sentences of the three use-cases available in BioXpress and compared the results in terms of correct sentence type and gene expression level. This means we consider our results correct when we extract the same gene and expression level as in the input data and when we correctly classify sentences as TypeA and TypeB with respect to the input data as well. 

Overall, our implementation correctly parse the 97\% of the sentences, these are sentences for which we are able to extract GDAs and that do not raise any exceptions during computation. Our system performance is shown in Table \ref{tab:results}. Results refer to all three use-cases and we consider the data provided in BioXpress as ground truth. The first two columns show the performance metrics when we run our implementation using the full DEXTER pipeline, i.e. using the PubTator annotations. The other two columns report the system performance when using the annotations provided by the input data instead of the PubTator annotations. As already mentioned, metrics are computed in terms of correct gene expression level and correct sentence type. 
\begin{table}[]
    \caption{Performance metrics. Results refer to all three use-cases and we consider the data provided in BioXpress as ground truth. The first two columns show the performance metrics when we run our implementation using the full DEXTER pipeline, i.e. using the PubTator annotations. The other two columns report the system performance when using the annotations provided by the input data instead of the PubTator annotations. Metrics are computed in terms of correct gene expression level and correct sentence type.}
    \label{tab:results}
    \begin{tabular}{lllll}
        \toprule
        & \multicolumn{2}{c}{Full Pipeline} & \multicolumn{2}{c}{Without PubTator Annotations}\\ 
        \cmidrule{2-5} 
        & \multicolumn{1}{c}{Expression Level} & \multicolumn{1}{c}{Sentence Type} & \multicolumn{1}{c}{Expression Level} & \multicolumn{1}{c}{Sentence Type} \\ 
        \midrule
        Accuracy  & 0.8436 & 0.9353 & 0.9293 & 0.9326\\
        Precision & 0.8436 & 0.9377 & 0.9296 & 0.9349\\
        Recall    & 0.8436 & 0.9352 & 0.9293 & 0.9326\\
        F1 score  & 0.8436 & 0.9359 & 0.9294 & 0.9332\\ 
        \bottomrule
    \end{tabular}
\end{table}
The results show that the system has been reproduced to a reasonable degree since the performance drop is negligible and mostly due to the use of SpaCy rather than the system implementation itself. In fact, after a deeper analysis we noticed several parse trees computed by SpaCy were different from the one returned by the CoreNLP library therefore most unparsed sentences can be attributed to it. Additionally, it is important to notice that PubTator is an automated tool based on a neural model that is periodically retrained therefore the version of PubTator we use to retrieve mentions is different from the one used in the original paper to compute the datasets we downloaded from BioXpress. As a consequence, missing mentions affect negatively the accuracy performance of the system as shown in Table \ref{tab:results}. In particular, expression level accuracy is 84\% when using PubTator mentions however if we use the mentions that have been already extracted by the original approach -- i.e., those using a different PubTator version -- the overall performance improves, especially as far as expression level is concerned where accuracy raise to 93\%. This means that most of the wrongly detected levels are due to the different version of PubTator and do not depend on the system implementation itself. Nevertheless, the dependency on PubTator limits the reproducibility of DEXTER and opens to some potential issues to apply it to other documents.

\section{Conclusions}
This work reproduces DEXTER, a system to automatically extract GDAs from biomedical abstracts.\cite{dexter} The goal is to provide a baseline for future works regarding RE, enabling researchers to test and compare their results. The system implementation parsed 97\% of the input sentences while discarded sentences are mostly due to missing PubTator annotations or problems related to Dependency Parsing. The system achieved an accuracy of 84\% on the gene expression level. Such value raises up to 93\% if input mentions are used instead of PubTator annotations, meaning that the gene expression level performance depends on the PubTator version not on the system implementation itself. Results in Table \ref{tab:results} demonstrate that the system is reproducible and it can be used as a benchmark for future works. 

In conclusion, implementing this system highlighted its limits. DEXTER is a rule-based model hence information are extracted based on patterns written to match sentences with a very specific syntactic structure, namely TypeA and TypeB sentences. If biomedical abstracts do not follow such structure, as for TypeC sentences, they are discarded. This is a huge limitation since sentences usually have a wide variety of syntactic forms and it could cause undesired loss of information. 
\begin{acknowledgments}
    This work was supported by the ExaMode Project, as a part of the European Union Horizon 2020 Program under grant 825292. We also sincerely thank Gianmaria Silvello and Stefano Marchesin whose suggestions helped in improving this study.
\end{acknowledgments}

\bibliography{bibliography}

\end{document}